# Road-network-based Rapid Geolocalization

Yongfei Li, Dongfang Yang*, Shicheng Wang, Hao He

Xi'an Research Institute of Hi-tech, China

## Abstract

It has always been a research hotspot to use geographic information to assist the navigation of unmanned aerial vehicles. In this paper, a road-network-based localization method is proposed. We match roads in the measurement images to the reference road vector map, and realize successful localization on areas as large as a whole city. The road network matching problem is treated as a point cloud registration problem under two-dimensional projective transformation, and solved under a hypothesise-and-test framework. To deal with the projective point cloud registration problem, a global projective invariant feature is proposed, which consists of two road intersections augmented with the information of their tangents. We call it two road intersections tuple. We deduce the closed-form solution for determining the alignment transformation from a pair of matching two road intersections tuples. In addition, we propose the necessary conditions for the tuples to match. This can reduce the candidate matching tuples, thus accelerating the search to a great extent. We test all the candidate matching tuples under a hypothesise-and-test framework to search for the best match. The experiments show that our method can localize the target area over an area of $400\ km^2$ within 1 second on a single cpu.

## 1. Introduction

With the increasingly wide use of unmanned aerial vehicles (UAVs), navigation for UAVs is becoming more and more critical. Generally speaking, there exist two kinds of methods for this task, recursive methods and global localization methods. The recursive methods accumulate measurements increasingly, such as angular velocity and acceleration from an Inertial Measurement Unit (IMU), or images from a camera. Such methods do not depend on external information and are very flexible. However, there is an inevitable defect for such methods that they all suffer from drift, because of their nature that they accumulate measurements throughout time or space. Moreover, they can only calculate the relative pose to a reference frame, which is usually the sensor coordination at the initial moment. To transform the relative pose to a certain navigation coordination, an initial alignment is usually needed.

Global localization methods are usually used to compensate the drift in recursive methods. Among these methods, the most widely used method is the satellite navigation guiding and positioning system, such as the GPS (Global Positioning System) or BDS (BeiDou Navigation Satellite System). Although they can provide continuous drift-free global localization, they are still faced with some challenges, like multipath effect, which can result in bad localization accuracy. Besides, they can be subject to obstructions from the aircraft's own body due to its attitude.

Considering that there are large numbers of features on the ground, such as roads, buildings, etc., which are rather robust across time, we propose a road network referenced localization method. We employ the road network information as reference data, because many works [1] [2][3] have been done in road segmentation, resulting in relatively accurate and robust results.

There are many advantages for localization using road network: first, reference road network are readily available from a Geographic Information System(GIS) data (such as road maps from OpenStreetMap); second, the presentation of road maps is highly compact, so only a small amount of storage is needed compared to aerial images or terrain contour maps; third, road networks are stable and they do not change much over time; finally, with the application of deep learning in road segmentation, we have achieved relatively accurate results. Although its application prospect is appealing, this task is quite demanding. Despite of high precision of the state-of-the-art road extraction methods, much noise still exists, which often causes great changes in road topology structure. And the reference road vector map may not be up-to-date, resulting in some differences between query aerial images and reference road network map. Furthermore, variations in scale, orientation and road width pose great challenge to register the road map estimated from the image captured by the onboard camera to reference road network. Under a homography transformation, which is the most general case, the length of line segment, the tangent, and the angle are not invariant, causing few features can be used to perform the matching. Because of this, many existing works[8][11][13][14][15][16] treat the transformation between the aerial image and the reference road network map as a similarity transformation, and extra attitude and altitude measurements from other sensors on the UAV must be used to transform the aerial image to the reference coordination. However, this may fail when large drift occurs. Finally, registration on a large area is very time consuming, and wise search strategy must be developed to enable a real-time application.

To address all those challenges, we treat the aerial-images-based localization problem as the 2D point cloud registration under a homography transformation. Different from registration under a Euclidean transformation or similarity transformation, most of the local features, such as tangent, angle and length of line, change with the transformation. But the order of contact: intersection and tangency stay unchanged[4]. So in this paper, road intersections are selected as the fundamental features. Moreover, road intersections can be extracted reliably using state-of-the-art road segment methods. Through road intersections contain few information when treated as individual, and thus less discriminative, the relative positions between them can fully constrain their correspondence. Inspired by this fact, we propose a global-feature-based registration method which works with pairs of intersections augmented with their tangents. We call the pairs of intersections 'two road intersections tuple', which is treated as a whole basic feature unit. It is shown that the homography transformation that aligns the detected road map and the reference road map can be determined in closed-form from a single pair of corresponding intersections tuples. And the rules to find the corresponding tuple candidates are designed to allow fast search. After that, all the putative tuple correspondences are tested in an hypothesise-and-test framework for global registration.

The rest of the paper is organized as follows: Section II presents a literature review; the detailed implementation of the proposed method is illustrated in Section III; Section IV contains experiments and analyses; following it are the conclusion and future research directions.

## 2. Related Work

Geographic information is appealing when applied to navigation, because of its stability over time and its ability to provide a global localization. Many efforts have been paid to localization using geographic information. According to different kinds of information they use, these methods can be divided into three classes: methods based on pre-collected aerial images, such as images from Google Earth or Google Street View, methods based on digital elevation maps, and methods based on GIS data.

Pre-collected aerial images, which are obtained via Google Earth, Google Street View and other methods, have been deployed for geo-referencing. In [5], visual odometry and image registration are used to augment the navigation system. Both the reference aerial images and images collected from onboard camera are converted to edge images. After a scale and rotation transformation, the two images are aligned to the same direction. And a template matching is performed to estimate the absolute pose for the UAV. Although the method is robust to changes in illumination, rotation and scale to a certain extent, the image registration relies on accurate estimation of attitude and altitude of the UAV, which limits the use of the method in practical application. Another approach [6] encodes the images with classification of the scene. Both UAV images and reference Google Map images are segmented into superpixels and then classified as grass, asphalt and house. Class histograms are constructed by selecting circular regions around a point, resulting in a rotation invariant descriptor. However, without using the rotation information, the method can only estimate the absolute translation between the UAV and the map. Moreover, the matching accuracy is poor in large homogeneous regions. In [7], Google Map images are used both for global localization and continuous navigation. The UAV position is initialized via correlation, after which the optical flow is used to predict its position in subsequent frames, and the onboard images are registered to Google Map with HOG features finally. In [8], road map and images around the road intersections are used to perform geolocalization. They use CNN to detect both roads and road intersections. Besides, the metric learning is used to find a descriptor for images around road intersections. They match roads intersections using that descriptor to find the candidate correspondences, and find accurate location with the geometric alignment of the two road maps. But matching road intersections using images around them cannot work, when there exist great differences between query images and reference images. This often occurs when the images are acquired on different weather conditions. There also exist some works deploying Google Street View images for UAV localization. In [9], searching in the Google Street View database is performed to provide the UAVs with a drift-free localization. To tackle with large viewpoint changes, some rendered views are generated. The Scale Invariant Feature Transform (SIFT) features are adopted to match onboard images to those rendered images. Nevertheless, SIFT requires intensive computation, making the method hard to achieve real-time application.

The digital elevation maps have long been used in navigation. [10] proposes an image-based terrain map matching method. In this method, a sparse elevation map is reconstructed with keypoints tracking and stereoscopic analysis. And then 3D surface interpolation is used to generate a dense, uniformly sampled elevation map. Instead of performing complex 3D registration, both the reference digital elevation map (DEM) and the recovered elevation map (REM) are converted to the 'cliff map' representation, resulting in a 2D shape matching problem. At last, the 2D shape matching problem is solved through critical points extraction, description and matching. In [11], 2D-3D tie-points, and geo-registered feature tracks are tightly incorporated into absolute geo-registered information. 2D-3D tie-points are established by finding feature correspondences to align an aerial video frame to a 2D geo-referenced image rendered from the 3D terrain database. Geo-registered feature tracks are generated by associating features across consecutive frames. But matching features from an onboard image to an image rendered from the 3D terrain database may be very challenging because of the great difference between the two kinds of images.

Also, many works focus on the use of GIS data in pose estimation of UAV. GIS uses spatio-temporal (space-time) localization as the key index variable for all other information. These GIS coordinates may represent other quantified systems of temporo-spatial reference (for example, highway mile-marker, etc.) [12]. Among all the information, roads and road intersections are the most widely used [13]-[16]. In [13], a road intersection extraction is firstly performed, and then all the road intersections within a certain radius is used to construct a 'start' centered at current road intersection. The length of every edge in the 'start' is calculated to form a descriptor. And the descriptor is applied on the search of corresponding road intersections which can be used to estimate the pose for the UAV. The method is performed under the hypothesis that lengths between intersections remain unchanged, which means the transformation is supposed to be a two-dimensional Euclidean transformation. This happens when the camera image plane and the altitude for the UAV are aligned to the reference coordinate, which is seldom the case in practical application. In addition, it is hard to match road intersections just using the distance information with other intersections. In [14] and [15], road intersection features are used. Instead of distinguishing road intersections only by local information (angles, length, etc.), a hypothesise-and-test framework is used, where two road intersections and their candidate correspondences are selected randomly to estimate a similarity transformation, and then all the left correspondences are tested on the calculated similarity transformation. To speed the search, local information, such as the angles of road junctions, is used to reduce candidate correspondences, and thus reducing the search space. In [16], they extract road segments in image sequences and search in a database created from a road network map for the best matches between the road database and the extracted road segments. Geometric hashing is used to retrieve a shortlist of matches which are ranked by a verification process. They prove that road map in a large enough area is unique to perform localization in an area as large as a whole city. But they suppose a similarity transformation between reference road map and query road map, which only happens when the camera image plane is parallel to the ground plane. Besides, they need minutes to perform a localization.

In summary, all those methods do not address the localization problem under a perspective transformation, which allows severe viewpoint changes between the query and the reference road vector map. Under a perspective transformation, many shape features, such as length of line segments, angle between line segments, cannot remain unchanged. So methods using such shape features will not work in such cases. Our main contributions are: 1) a global projective invariant feature named two road intersections tuple is proposed, and the necessary conditions to find matching tuple is studied; 2) our method utilizes road networks as appearance invariant features to localize an aerial image over a large area, and experiments show that our method can perform a successful localization within seconds, or even within tens of milliseconds when special road intersections exist in query image.

## 3. Road-intersections-based Localization

In this chapter, we will introduce road-intersections-based localization method in detail. We start by describing the road intersections detection method, and defining the perspective invariant descriptor for an individual intersection. Following that, we introduce our two road intersections tuple feature and the necessary conditions for a two intersections tuple feature to be a correspondence of another. Finally, the two road intersections tuple feature based localization method will be presented fully.

### 3.1 Perspective invariant feature for intersection description

Our road intersections detection is based on the road segmentation. Many algorithms have been developed for road segmentation. Any method that can detect road effectively can be used in our localization method. In this article, we use method proposed in [17].

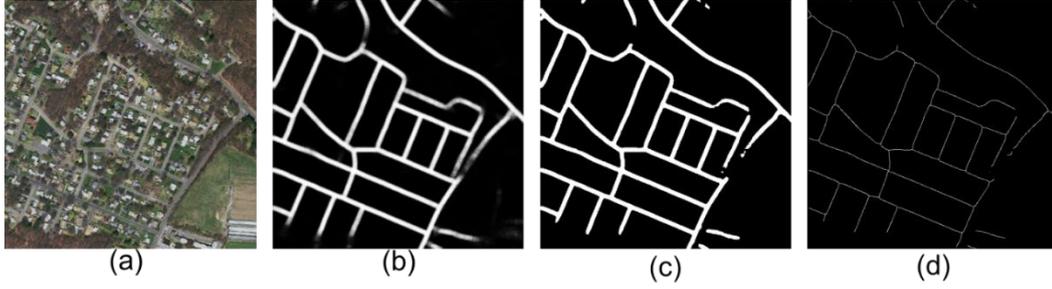

*Figure 1. Road intersections detection and description. (a)Query aerial image. (b)Road segmentation result. (c)Binary road image. (d) Skeletonized road image.*

As shown in Figure 1, the morphological filtering is conducted on the segmented image to generate a clean road binary image after road segmentation. And then, the binary road image is skeletonized. After that, all the points with more than two neighbor points are treated as road intersection. In practice, there often exist more than one such kind of points in a road intersection, so we merge points within a certain radius to form a road intersection. At last, we extract $n$ interconnecting pixels on the road skeletons to find all the branches for the road intersection.

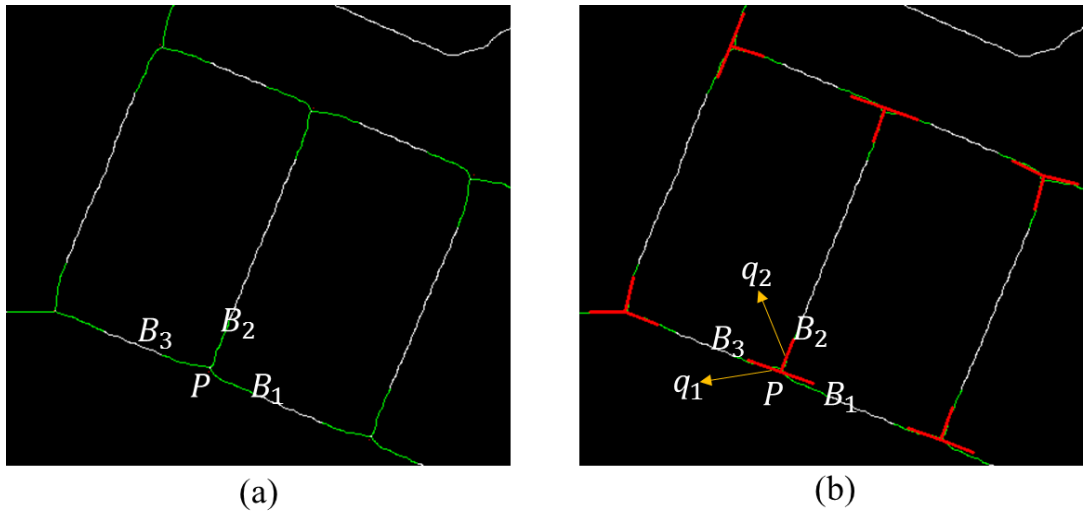

*Figure 2. Road intersection branches segmentation and tangents estimation. There are 3 branches for intersection $P$, and 3 lines are fitted using the points of the 3 branches respectively. We merged lines between which the angle is smaller than a certain threshold. Thus, there are 2 tangents for intersection $P$. The branches are drawn in green, while the tangents are in red.*

As shown in Figure 2, we estimate the tangents for the road intersection $P$ using the point sets in the extracted road intersection branches $P_{Bi} = \{p_{ij}\}(i = 1, 2, ..., n)$. For each branch point set, a line fitting is conducted to get the tangent line $l_i$ and tangent vector $\mathbf{q}_i$. Because the road intersection detection method above cannot estimate the intersection center accurately, we perform a fine tuning for the intersection center using all the branch lines of it:

$$\mathbf{l}_i^T \mathbf{x} = 0, i = 1, 2, ..., n \tag{1}$$

Considering that the tangent lines fitted by different branch point sets may be collinear and should be treated as one tangent line, we merge tangent lines whose angles are below a certain threshold:

$$\langle \mathbf{q}_i, \mathbf{q}_j \rangle < \delta, \quad (2)$$

where $\langle \mathbf{q}_i, \mathbf{q}_j \rangle$ is the angle between tangent line $\mathbf{q}_i$ and $\mathbf{q}_j$, and can be calculated as:

$$\langle \mathbf{q}_i, \mathbf{q}_j \rangle = \frac{\mathbf{q}_i \cdot \mathbf{q}_j}{|\mathbf{q}_i||\mathbf{q}_j|}, \quad i,j = 1,2,...,n \quad (3)$$

Because the intersection for lines and tangent line for curve remain invariant under a 2D perspective transformation, we choose the following simple features to describe an individual road intersection: the number of branches in the road intersection $N_B$, and the number of tangent lines in the road intersection $N_q$.

In practice, there exist only a few possible values for $N_B$ and $N_q$ (typically, $N_B=3,4,5$ and $N_q=2,3,4,5$). So it is unrealistic to perform road intersection matching with such simple descriptor. But with the relative position between road intersections, it is usually distinguishable enough. Hence, we propose the two road intersections tuple feature.

### 3.2 Two road intersections tuple feature and matching rules

A homography transformation can be estimated uniquely with four independent point correspondences or four independent line correspondences. Here 'independent' means any three of the given four points are not collinear, or any three of the given four lines are not concurrent [4]. Since there are at least two tangents at one intersection, two matching intersection pairs and their corresponding tangents can determine a homography transformation uniquely. When there are more than two tangent lines at the intersection, the additional tangent lines can provide redundant constraint which can be used to provide constraint for the two road intersections tuple to match, resulting in a reduce in search space.

1. Cross ratio descriptor

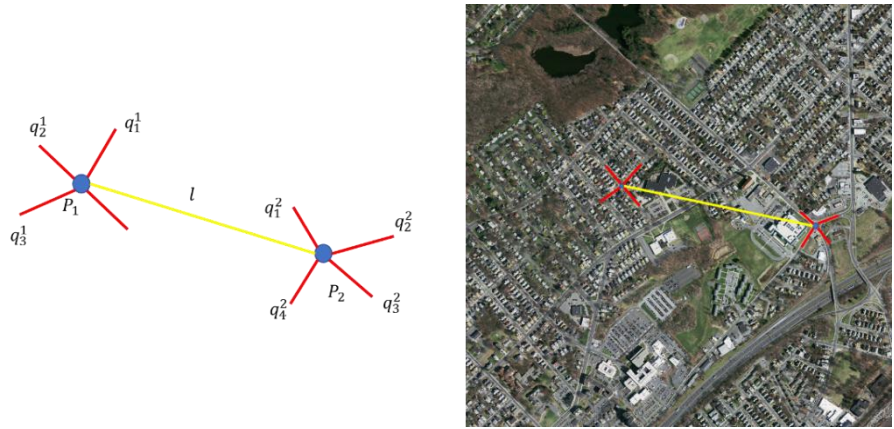

*Figure 3. Two road intersections tuple feature.*

As shown in figure Figure 3, let $P_1, P_2$ be a pair of intersections, $Q_1, Q_2$ are their corresponding tangent line sets, and $l$ is the connection between $P_1, P_2$. And all lines passing through intersection $P_i (i=1,2)$ can be presented as $M_i = Q_i \cup l (i = 1,2)$. Let $N_{M_i}$ be the number of elements in $M_i$. When $N_{M_i}$ is no less than 4, there exist at least four lines passing through the intersection, and thus cross ratios can be estimated using these lines. Supposing there are $N_{M_i}$ lines at the intersection $P_i$, $C_{N_{M_i}}^4$ cross ratios can be determined. These cross ratios contain the relative position information for the two road intersections tuple, and we call them 'cross ratio descriptor' in this paper:

$$\mathbf{d}_c = \left[ c_1^1, c_2^1, ..., c_{N_1}^1, c_1^2, c_2^2, ..., c_{N_2}^2 \right]^T, N_i = C_{N_{M_i}}^4 \ (i = 1, 2), \quad (4)$$

where $c_j^k$ is the $j$th cross ratio for $k$th intersection.

The cross ratio is invariant under a homography transformation, so the necessary condition for a two road intersections tuple to be a match of anther is:

$$\mathbf{d}_c = \mathbf{d}_c', \quad (5)$$

Considering the noise in tangent lines estimation and cross ratios estimation, we propose a relaxation condition as:

$$\frac{|\mathbf{d}_c - \mathbf{d}_c'|_i}{|\mathbf{d}_c|_i} < \delta, i = 1, 2, ..., C_{N_M}^4, \quad (6)$$

where $|\cdot|_i$ are the $i$th dimension for cross ratio descriptor, and $\delta$ is the relative error threshold.

2. Matching rules for two road intersections tuple

Even for a matching two road intersections tuple pair, the correspondence between their tangent lines are not unique. Let $P_i' - Q_i'$ be the corresponding two road intersections tuple for $P_i - Q_i$, and then the connections of the two intersections centers $l$ and $l'$ are surely to be corresponding. Therefore, there exist only two kinds of correspondence for tangent set $M_i$ and $M_i'$: all the tangents are matched to each other clockwise, or one clockwise while the other counterclockwise.

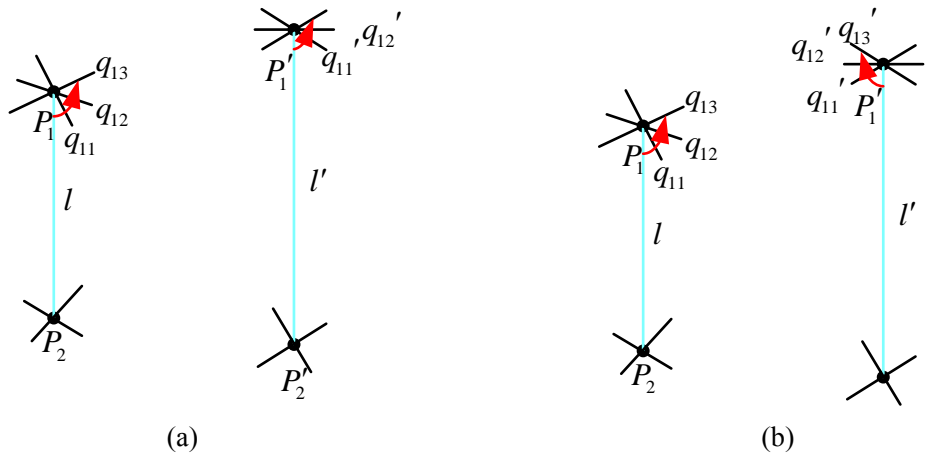

(a) (b)

*Figure 4. All possible correspondences of tangent lines for a matching two road intersections tuple pair. Because the connections of the two intersections centers are surely to be corresponding, there exist only two kinds of correspondence for tangent lines at one matching intersection pair: (a) all the tangents are matched to each other clockwise, or (b) one clockwise while the other*

*counterclockwise. Thus, there exist four possible correspondences for tangent lines of a matching two road intersections tuple pair.*

Thus, the matching rules for two intersections tuple are:

1) $\mathbf{d}_b = \mathbf{d}'_b$, where $\mathbf{d}_b = \left[ N_B(P_1), N_B(P_2), N_q(P_1), N_q(P_2) \right]^T$ are the branch descriptor;

2) $\dfrac{\left|\mathbf{d}_c^{++} - \mathbf{d}_c'^{++}\right|_i}{\left|\mathbf{d}_c^{++}\right|_i} < \delta, i=1,2,...,C_{N_M}^4$, or $\dfrac{\left|\mathbf{d}_c^{++} - \mathbf{d}_c'^{+-}\right|_i}{\left|\mathbf{d}_c^{++}\right|_i} < \delta, i=1,2,...,C_{N_M}^4$,

or $\dfrac{\left|\mathbf{d}_c^{++} - \mathbf{d}_c'^{-+}\right|_i}{\left|\mathbf{d}_c^{++}\right|_i} < \delta, i=1,2,...,C_{N_M}^4$, or $\dfrac{\left|\mathbf{d}_c^{++} - \mathbf{d}_c'^{--}\right|_i}{\left|\mathbf{d}_c^{++}\right|_i} < \delta, i=1,2,...,C_{N_M}^4$,

where + means the cross ratios are estimated when the tangent lines are sorted clockwise, meanwhile − presents cross ratios are estimated when the tangent lines are sorted counterclockwise.

### 3.3 Two-road-intersections-tuple-based localization

At this point, given the query image, we are able to extract all the two intersections tuples in it and find all the matching candidate tuples for them in the reference road vector map. But usually there are many matching candidate tuples for a two intersections tuple, and further verification should be conducted to find the unique correspondence. A corresponding two intersections tuple pair can determine a homography transformation uniquely, which can be used to verify the correspondence together with other intersections. We perform this in an RANSAC frame. The whole localization algorithm is shown in Alg.1:

---

**Algorithm 1 The Two-road-intersections-tuple-based Location Algorithm.**

**Input:** $P_q$-intersections in query image; $P_r$-intersections in reference image;

$\lambda, \lambda_1$-confidence; $\delta$-inlier rate threshold.

**Output:** $\mathbf{H}$-homography matrix; $MP$-matching pairs;

1: $\theta^*_{inlier} \leftarrow 0$.

2: **for** i=1→$l(\lambda, \lambda_1, p_{query})$ **do**

3:     $S^i_{P_q} \leftarrow$ Draw a two-intersections sample in query image.

4:     $P^i_{r-candidate} \leftarrow$ Find all the two-intersections feature candidates passing through the matching check in reference map.

5:     **for** j=1→$k(\lambda_1, p_{query})$ **do**

6:         $S^{i,j}_{P_r} \leftarrow$ Draw a two-intersections sample in $P^i_{r-candidate}$.

7:         $\mathbf{H}_{ij} \leftarrow$ Estimation the homography matrix with $S^i_{P_q}$ and $S^{i,j}_{P_r}$

8:         $\theta^{ij}_{inlier}, MP_{ij} \leftarrow$ Compute the inlier rate and find matching intersections.

9:         **if** $\theta^{ij}_{inlier} > \theta^*_{inlier}$ **then**

10:             $\mathbf{H}^*, MP^*, \theta^*_{inlier} \leftarrow \mathbf{H}_{ij}, MP_{ij}, \theta^{ij}_{inlier}$

11:     **if** $\theta^*_{inlier} > \delta$ **then**

12:         return.

13: Fail to locate. **End.**

---

**Sampling in query image:** Given a query image, we extract all the two road intersections tuples first. Since at least four lines among which none three lines should pass through a common point are needed to determine a homography transformation, the tangent lines for one intersection in the tuple should not pass through the other intersection. This is done by:

$$\langle q, l \rangle > \delta, \tag{7}$$

where $q$ presents the tangent line at the intersection, and $l$ is the connection between two intersections.

Usually the numbers for different types of two intersections tuple (we treat tuples with the same number of tangent lines and same number of branches as the same type) vary a lot in the reference map. The fewer tuples exist in one type in the reference map, the fewer candidate tuples there are if we sample such type of tuple in the query image. This can lead to a reduction in the search space. Thus, the priority for different types of tuple in query image are determined by the numbers of tuple for that type in reference map: the fewer tuples are in that type in the reference map, the higher the sampling priority for that type is.

**Sampling in reference map**: the extraction of two intersections tuples for reference image are done offline. This is conducted with the same way as that for query image with only one small difference. Considering it is impossible to observe simultaneously two intersections that are far away from each other, only two intersections within a certain distance can form a two intersections tuple:

$$|P_1 - P_2| < d_{max} \tag{8}$$

The distance threshold $d_{max}$ can be estimated with the approximate flight altitude of the UAV and the visual angle of the camera.

In practice, there are only a few dozen types of two intersections tuple, since all possible values for descriptor $\mathbf{d}_b = [N_B(P_1), N_B(P_2), N_q(P_1), N_q(P_2)]^T$ are finite and discrete. To enable a fast search, we save all the two road intersections tuples in the reference map in a hash table. The descriptor $\mathbf{d}_b = [N_B(P_1), N_B(P_2), N_q(P_1), N_q(P_2)]^T$ is put as the key for the hash table, and tuples with the same descriptor are saved in the value with the same key. Thus, every value in the hash table contains tuples with the same number of tangent line and the same number of branch. Then, we calculate the cross ratio descriptors $\mathbf{d}'_c$ for all tuples in one value. Since the number of tangent line of the two intersections tuple in a certain value is the same, the dimension of cross ratio descriptors for tuples in the same value are also the same. Following that, we build a search tree with the cross ratio descriptors $\mathbf{d}'_c$ for every kind of tuples, resulting in a group of cross ratio search tree: $Tree_c = \{tree_c(\mathbf{d}_b) | \mathbf{d}_b \in \mathbf{D}_b\}$, where $\mathbf{D}_b$ are all the possible values for $\mathbf{d}_b = [N_B(P_1), N_B(P_2), N_q(P_1), N_q(P_2)]^T$.

For a given sample in query image, all the candidate matching tuples in reference map are found using the rules proposed in section 3.2. And we choose one tuple randomly from them as the sample of reference map.

**Model estimation**: the homography transformation is estimated using all the tangent line correspondence as:

$$\mathbf{l}' = \mathbf{H}^{-T} \mathbf{l} \tag{9}$$

with the gold standard algorithm [4].

**Model evaluating**: the inlier ratio in query image is adopted to evaluate the homography transformation model estimated by the sampled two intersections tuple corresponding pair. An intersection in the query image is treated as an inner point under a certain homography transformation, when there exists one intersection close enough to the transformed center for the given intersection in the reference map, and they must belong to the same type. So the inlier ratio can be calculated as:

$$\theta_{inlier} = \frac{N(\{\mathbf{x} \mid \left| [\mathbf{Hx}]_{1,2}/[\mathbf{Hx}]_{1,3} - \mathbf{x}'_{1,2}/\mathbf{x}'_3 \right| < \delta\})}{N_{sum}} \qquad (10)$$

where $[\bullet]_{1,2}, [\bullet]_3$ are the first two dimensions and the third dimension of the homogeneous coordinate $\mathbf{x}$ and $\mathbf{x}'$ respectively, and $\mathbf{H}$ is the estimated homography matrix.

**Termination condition**: a high enough inlier ratio means enough intersections have found their corresponding intersections with the same type in the reference map. Later, we will show in our experiments that enough intersections correspondences are able to determine the location uniquely. So a localization is regarded as a success when $\theta_{inlier} > \delta_{inlier}$.

There are also cases when no matching location is in the reference map for a certain query image. We deduce the conditions for a localization considered a failure. There are two cases when a localization fails. The one is when there exists the corresponding location for the query image, but we fail to find it; the other is when there does not exist a corresponding location for the query image. We should try our best to avoid the first case. Usually there are two reasons for the first case. In one case, the corresponding tuple in the reference map are not sampled for a given two road intersections tuple $P_1 - P_2$ in the query image, and in the other case, the corresponding tuple does not exist in the reference map for a given two intersections tuple $P_1 - P_2$ in the query image. This happens when big errors occur in road segmentation or/and estimation for tangent lines.

Suppose there are $N_{match}$ candidate matching tuples in the reference map for a given two intersections tuple $P_1 - P_2$ in the query image, the probability of failing to find its corresponding tuple after $k$ sampling in reference map is:

$$P_{fail-reference} = \left(1 - 1/N_{match}\right)^k. \qquad (11)$$

Suppose the probability that there exists the corresponding two intersections tuple in reference map for a given tuple in query image is $P_{query}$, the probability of failing to localize after $l$ sampling tries in query image is:

$$P_{fail-sum} = \left(1 - P_{query} + P_{query} P_{fail-reference}\right)^l. \qquad (12)$$

So when we require the probability that the first case for a failed localization happens is less than $\lambda$, the minimum sampling time is:

$$\langle k,l \rangle = \arg_{k,l}(P_{fail-sum} < \lambda). \tag{13}$$

To solve equation (13), we introduce an additional constraint:

$$P_{fail-reference} < \lambda_1. \tag{14}$$

It means the probability that we fail to find the corresponding tuple when it exists actually should be less than $\lambda_1$. Thus, for $ith$ sampling in the query image, the minimum sampling time required in reference map is:

$$k_i > \frac{\log \lambda_1}{\log\left(1 - 1/N^i_{match}\right)}. \tag{15}$$

Furthermore, the minimum sampling time required in query image is:

$$l > \frac{\log \lambda}{\log\left(1 - P_{query} + P_{query}\lambda_1\right)}. \tag{16}$$

So the total minimum sampling time required is: $N_{sum} = \sum_{i=1}^{l} k_i$. This means if we sample $N_{sum}$ times and fail to find the location for query image, we can regard that there exist no matching area in the reference map for the given query image with a confidence of $1-\lambda$, hence we can stop the search.

## 4. Experimental results

### 4.1 Dataset

We used the Massachusetts roads dataset[18] to test our algorithm. The dataset contains aerial images covering more than $2600\,km^2$ in Massachusetts. All the images are RGB images with size $1500 \times 1500$, and the resolution for these images is 1 m/pixel. In order to improve the generalization performance of the training model, manual occlusions are added to some images in the train set. Those images were removed from the dataset for our experiments. All images are named by the coordinate of image center, which is expressed in the EPSG:26846[1]. This provides the ground truth for the image location. In addition to the aerial images, a road vector map for the corresponding area is also provided in the form of an ERS Shapefile file.

Since all the aerial images are orthorectified, four random perspective transformations were conducted to every image in the dataset to generate 4 more images. They were used to test the effectiveness of our algorithm when there is a homography transformation between the query image and the reference map.

The whole area in the vector map is as large as approximate $60000\,km^2$, we split the region into subareas with size $20\,km \times 20\,km$. The road vector map for every area was extracted, and transformed to binary images with resolution of 1 m/pixel. So the image size was $20000 \times 20000$. And all the aerial images were classified into the corresponding subareas according to their center coordinates.

---

1 http://spatialreference.org/ref/epsg/26846/

## 4.2 Implementation details

**Reference map:** There are usually thousands of intersections in a reference image. To extract all the intersections and calculate their tangent lines are a little time-consuming (a few minutes), but it can be conducted offline. We extracted the intersections with method proposed in section 3.1, and 15 connected pixels were picked for every branch at one intersection. In order to estimate tangent lines accurately, any intersection with branch that contained less than 15 pixels, or whose average error of tangent fitting is greater than 0.5 pixel, was discarded. The intersections were then saved in a file for further use.

**Result evaluation:** A homography matrix can be calculated using our method, with which the image center in the reference coordination can be estimated. This was compared to ground truth to validate whether the localization was correct. And our experiments showed that the aerial image coincided well with the reference road image visually, when the estimated coordinate was close enough to the ground truth.

Our implementation was written in C++, single threaded. The experiments run on an Intel Xeon E5-2620v2 processor with 12 cores.

## 4.3 Experiments using synthetic data

We conducted experiments on synthetic data to investigate whether the intersections together with their tangent lines can determine the location of query image over a whole city uniquely. And they were also used to test the influence of parameters on the performance of the algorithm. We started by studying the minimum number of intersections needed to localize uniquely. And then the effect of inlier ratio on localization accuracy was analyzed. Finally, we discussed the efficiency of the algorithm.

We selected two search areas in the Massachusetts, one in Worcester, the other in Boston, both with an area of $10 \times 10 km^2$. To generate the query images, we added random homography transformations on the whole area, and then extract areas with size $d_r$ randomly. At the same time, the homography transformations and the area centers were recorded to provide a ground truth.

To test how many intersections are enough to localize uniquely, an intersection subset with $N_r$ intersections in every query image was selected randomly. We calculated the $precision = TP / (TP + FP)$ where $TP$ is the number of successful and true localization, and $FP$ is the number of successful but false localization. In figure Figure 5, the precision was plotted over the number of selected intersections for the two cities with different scene sizes $d_r \times d_r$.

The precision values derived from our experiment indicate, as shown in figure Figure 5, that enough intersections together with their local geometry information are unique. For both cities, we achieve a precision higher than 0.9 with no less than 30 intersections used in the localization. And our test also shows that the size of scene has little impact on the precision, and a high precision can be got as long as enough intersections are observed in the scene.

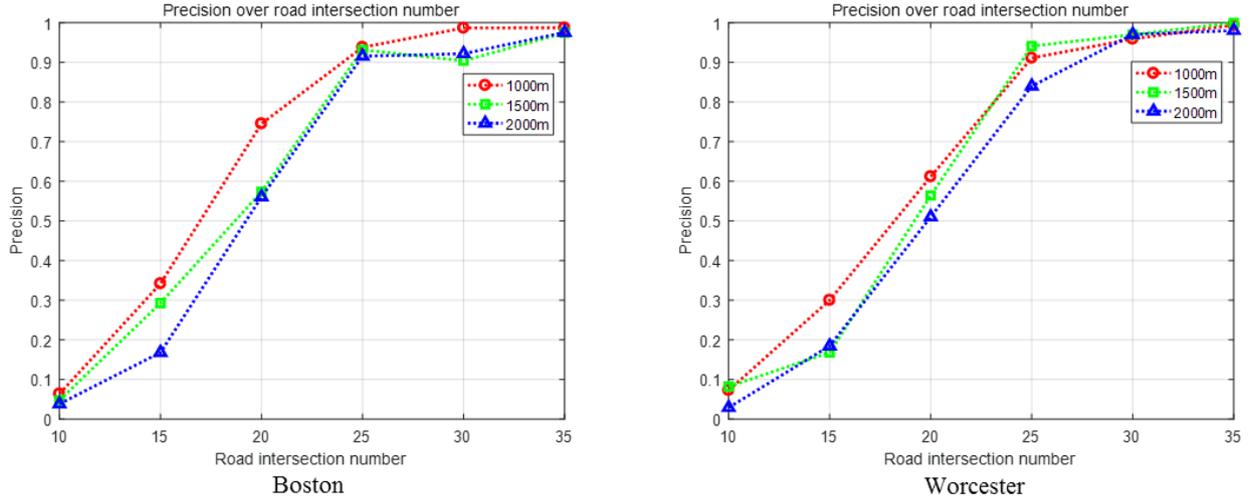

Figure 5. Precision over road intersection number with different scene sizes.

The inlier ratio is used to evaluate whether the localization is successful in our method. Thus, it has a big influence on the precision and the recall. We studied the impact of inlier ratio threshold. In this test, the scene size was set to $1.5 \times 1.5 km^2$, and the minimum number of intersections need to perform the localization was set to 30. The results for both cities are shown in figure Figure 6:

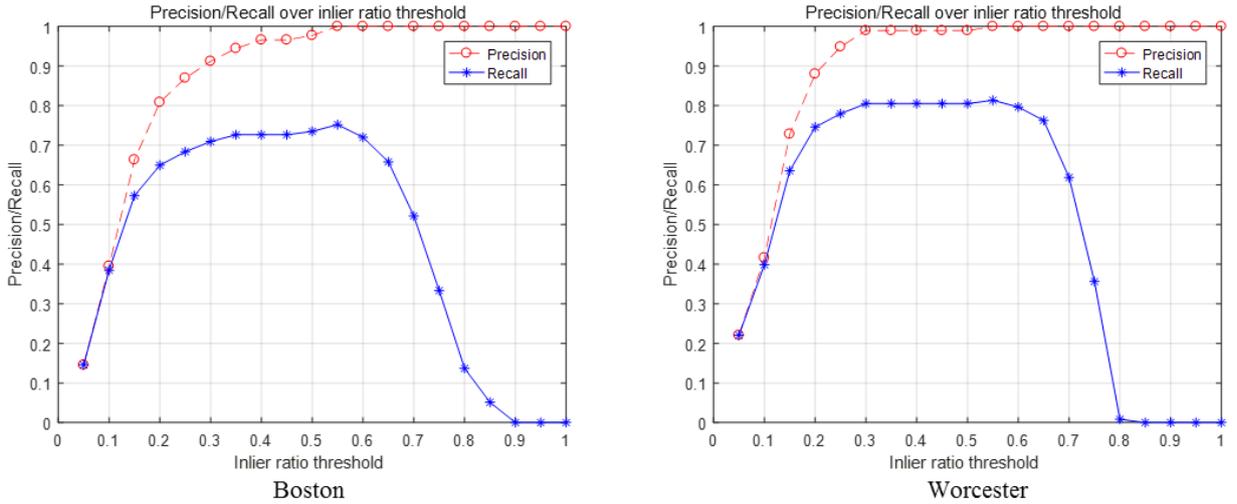

Figure 6. Precision and recall over inlier ratio threshold.

We can see that both the precision and the recall are low when the inlier ratio threshold are set to a small value. When a small inlier ratio threshold is chosen, the algorithm tends to find a false area and stop its search early, leading to low precision and recall. When the inlier ratio threshold is set to a large value, we achieve a high precision at the cost of a low recall. The reason may be that a high inlier ratio threshold is too strict to accept some correct localizations. And we obtain 100% precision and recall higher than 70% when the inlier ratio threshold is set to a reasonable value in both cities.

The time needed for performing a localization varies from tens of milliseconds to tens of seconds, depending on the intersections observed. There are usually thousands of intersections, and millions of two road intersections tuples in the search area. But the distribution of tuple type is extremely uneven. The distributions of tuple type for Boston and Worcester are shown in figure Figure 7:

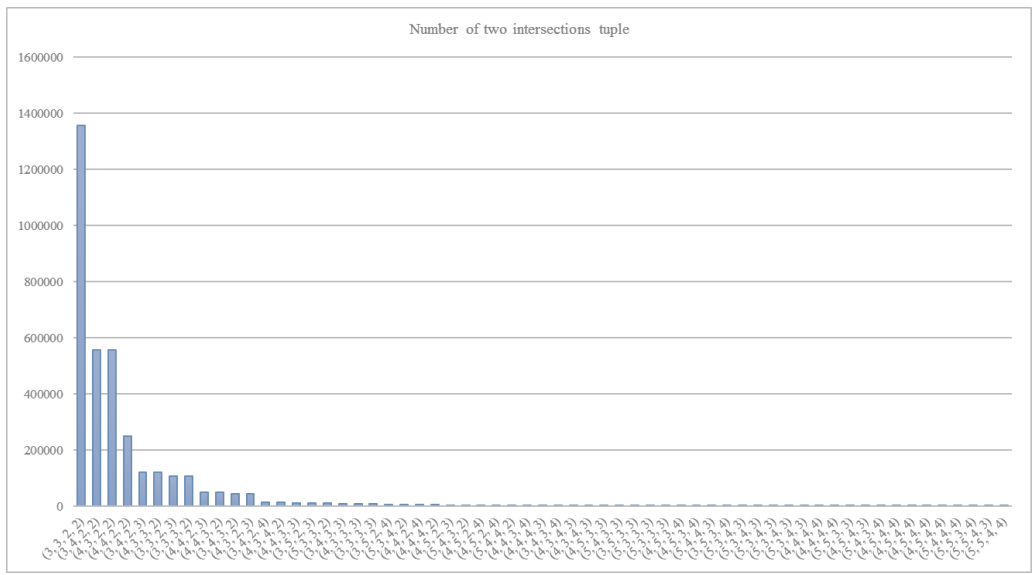

Boston

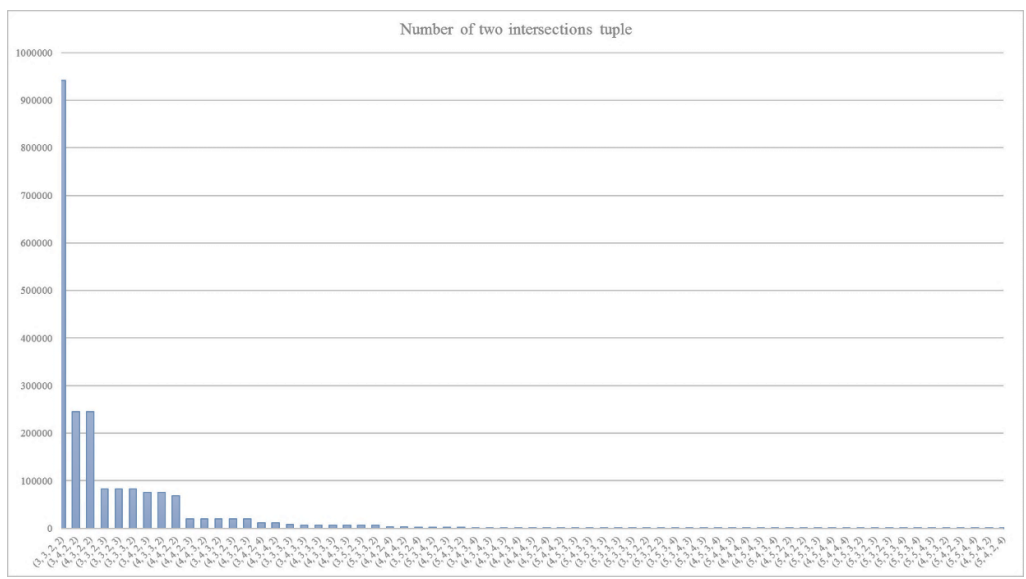

Worcester

*Figure 7. Distribution of different types of two road intersections tuples.*

The distribution of tuple type results in the big difference in run time. When there exist 'special' two intersections tuples in the query image (a tuple with a type whose tuple number is relatively small in the search area is called 'special tuple'), the search runs very fast. This is because there exist few candidate tuples in the search area for a 'special' tuple. The distributions of run time in both cities are shown in figure Figure 8:

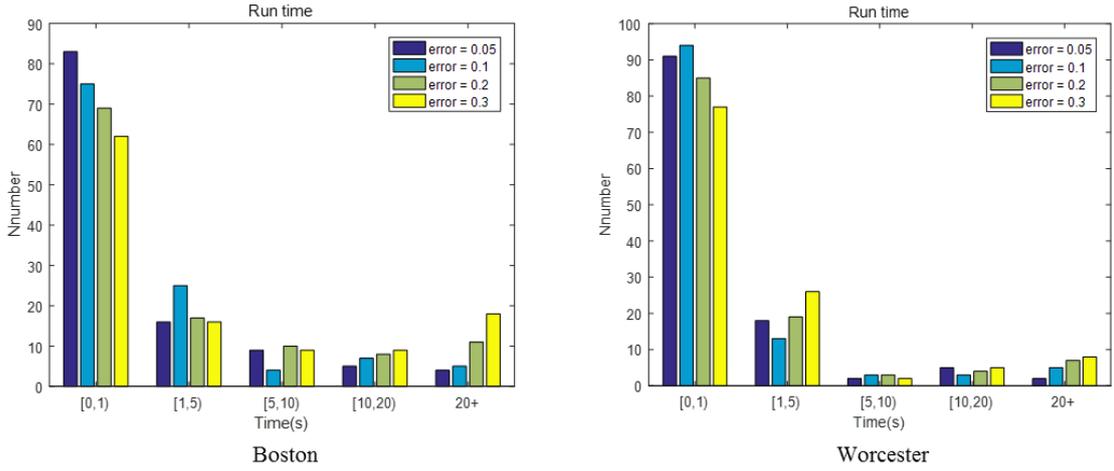

*Figure 8. Run time in Boston dataset with scene size of $1.5\times1.5km^2$ under different cross ratio relative errors.*

Here we tested the influence of cross ratio relative error threshold on run time. A smaller cross ratio relative error threshold can refuse more wrong candidate tuples, leading to faster search. However, because of the noise in cross ratio estimation, true candidate tuples may also be refused when the cross ratio relative error threshold is too small, leading to a decrease in recall. Figure 8 shows that it takes less than 1 second for a localization in most cases (more than 60%), which is much faster than that reported in [16] (several minutes).

## 4.4 Experiment on aerial image dataset

We performed localization experiments on subareas areas in Massachusetts Roads. If there are too few images in a certain subarea, we cannot calculate the precision accurately considering the randomness. Hence, all subareas with images less than 20 were not used in our experiment. At last, we obtained 17 subareas, and their parameters are shown in table Table 1:

| AREA ID | Area Center (m) | Image Number |
|---|---|---|
| 1 | 104175.9241, 889718.5602 | 39 |
| 2 | 104175.9241, 869718.5602 | 91 |
| 3 | 124175.9241, 869718.5602 | 71 |
| 4 | 164175.9241, 889718.5602 | 75 |
| 5 | 184175.9241, 889718.5602 | 77 |
| 6 | 184175.9241, 869718.5602 | 73 |
| 7 | 204175.9241, 889718.5602 | 82 |
| 8 | 224175.9241, 949718.5602 | 35 |
| 9 | 224175.9241, 909718.5602 | 83 |
| 10 | 224175.9241, 889718.5602 | 113 |
| 11 | 244175.9241, 929718.5602 | 71 |
| 12 | 244175.9241, 909718.5602 | 88 |
| 13 | 244175.9241, 889718.5602 | 70 |
| 14 | 244175.9241, 869718.5602 | 21 |
| 15 | 244175.9241, 849718.5602 | 39 |
| 16 | 264175.9241, 929718.5602 | 26 |
| 17 | 264175.9241, 869718.5602 | 57 |

*Table 1. Number of query images in different areas.*

In all the experiments, the minimum number of intersections needed was set to 30, and the inlier ratio threshold was set to 0.5. The cross ratio relative error was set to 0.3, considering the large noise in road segmentation. We computed the precision and recall in every subarea, and the results are shown in table Table 2:

| AREA ID | Precision（%） | Recall（%） |
|---|---|---|
| 1 | 80.2 | 70.6 |
| 2 | 76.9 | 68.9 |
| 3 | 92.3 | 68.8 |
| 4 | 86.7 | 67.1 |
| 5 | 80.0 | 69.4 |
| 6 | 100.0 | 88.9 |
| 7 | 87.5 | 74.5 |
| 8 | 81.3 | 68.4 |
| 9 | 74.6 | 54.9 |
| 10 | 72.6 | 58.2 |
| 11 | 96.3 | 65.0 |
| 12 | 76.9 | 61.9 |
| 13 | 72.7 | 57.1 |
| 14 | 86.7 | 50.0 |
| 15 | 100.0 | 70.0 |
| 16 | 100.0 | 100.0 |
| 17 | 87.5 | 70.0 |

*Table 2. Precision and recall in different areas.*

Compared to the experiments on synthetic data, both the precision and recall reduce slightly. This is mainly because there exists large noise in road segmentation result, leading to inexact estimation of tangent lines of intersections. With an enhanced road intersection segmentation method, we may expect a better performance.

We also show some successful localizations in figure Figure 9:

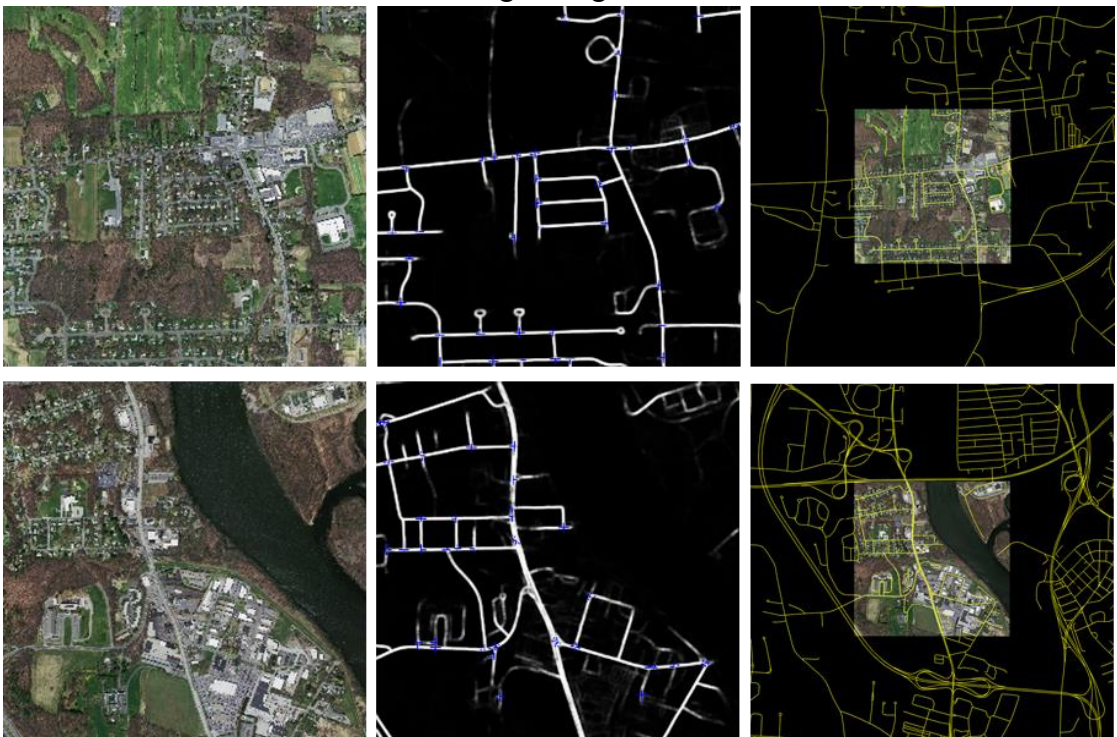

*Figure 9. Localization result. The original aerial images are shown in first column, the road segmentation result and road intersections detection result are in the middle column, and the localization result are shown in the last column. The reference road vector map is in yellow. We draw the aerial images on the reference road vector map with the estimated homography transformation.*

*A padding with size 1000m is added to show the surrounding reference road map.*

The query images are aligned in direction in experiments above, which means there exist only translation between the query image and the reference road vector map. To test the performance of our method on query images with perspective transformation, we added 4 random perspective transformations for every aerial image to generate 4 more query images. And then we performed experiments on these generated images, the result is shown in table Table 3:

| AREA ID | Precision (%) | Recall (%) |
| --- | --- | --- |
| 1 | 86.3 | 73.8 |
| 2 | 74.5 | 53.9 |
| 3 | 96.9 | 92.8 |
| 4 | 85.0 | 62.3 |
| 5 | 83.3 | 71.5 |
| 6 | 100.0 | 92.9 |
| 7 | 92.1 | 78.8 |
| 8 | 84.6 | 62.4 |
| 9 | 75.4 | 55.9 |
| 10 | 76.9 | 62.6 |
| 11 | 98.4 | 54.0 |
| 12 | 81.4 | 66.3 |
| 13 | 74.4 | 57.8 |
| 14 | 76.5 | 54.2 |
| 15 | 100.0 | 58.2 |
| 16 | 85.7 | 85.7 |
| 17 | 88.9 | 72.7 |

*Table 3. Precision and recall for different areas using aerial images with manual perspective transformations.*

As shown in table Table 3, our algorithm performs well when perspective transformation exists between query images and reference vector map. Some successful localizations are shown in figure Figure 10:

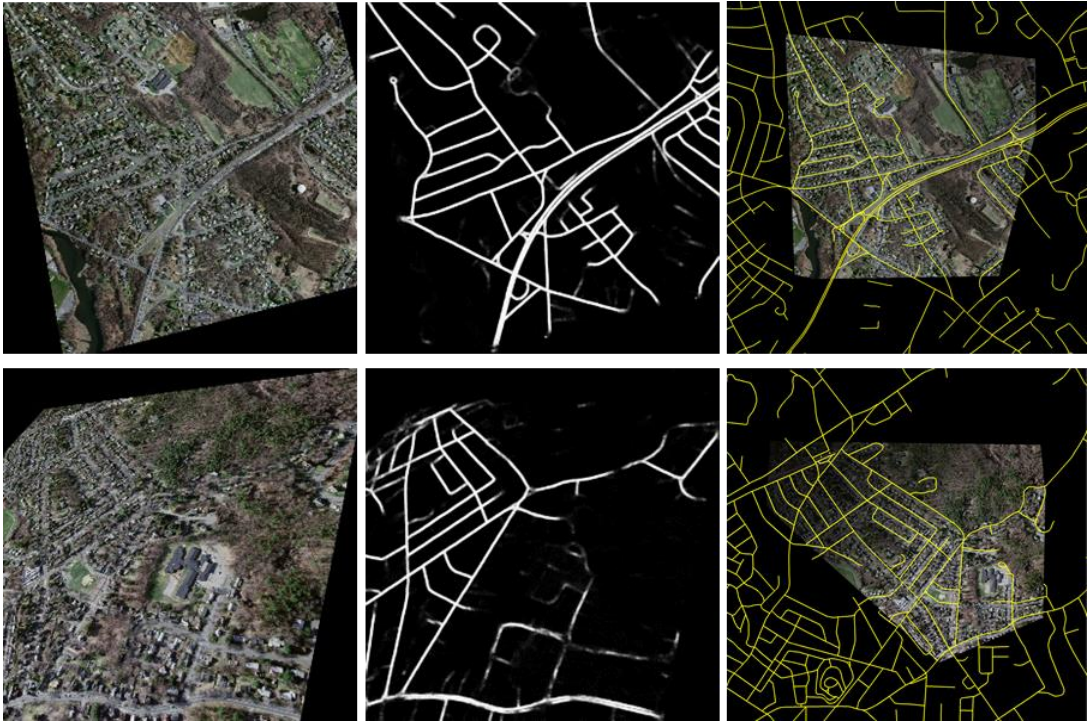

*Figure 10. Localization result on images with manual perspective transformations.*

# 5. Conclusion and outlook

In this paper, we address the problem of geolocalizing an aerial image with road in a large area. We use only road vector map as reference which is easily accessible and highly compact. We have demonstrated the effectiveness of our method in experiments on synthetic data and real aerial images. In our current work, we require there are enough intersections in the aerial images, which limits the use of our method for UAVs flying in a relative low altitude. In fact, we use only the relative positions between intersections in the localization, so we can accumulate the observation of intersections over image sequences to overcome the limitation, which is left for our later work.

# 6. Acknowledgement

This paper is supported by the National Natural Science Foundation of China (Grant No. 61673017，61403398), and the Natural Science Foundation of Shaanxi Province(Grant No. 2017JM6077, 2018ZDXM-GY-039).